\pdfoutput=1
\documentclass{article}
\usepackage{microtype}
\usepackage{graphicx}
\graphicspath{{figures/}}
\usepackage{booktabs}
\usepackage{multirow}
\usepackage{multicol}
\usepackage{hyperref}
\usepackage{amsmath}
\usepackage{amssymb}
\usepackage{mathtools}
\usepackage{amsthm}
\usepackage[capitalize,noabbrev]{cleveref}
\usepackage{xcolor}

\usepackage[preprint]{preprint}

\preprinttitlerunning{UpLoop: Training Looped Transformers via Upcycling}

\begin{document}

\twocolumn[
\preprinttitle{UpLoop: Training Looped Transformers via Upcycling at No Extra Pretraining Compute}

\preprintsetsymbol{equal}{*}

\preprintsetsymbol{corresponding}{\smash{\ensuremath{\dagger}}}
\begin{preprintauthorlist}
\preprintauthor{Cheng Liang}{equal,sjtu}
\preprintauthor{Tengfei Zhang}{equal,ustc}
\preprintauthor{Ya Zhang}{sjtu}
\preprintauthor{Yanfeng Wang}{sjtu}
\preprintauthor{Zhouhan Lin}{sjtu,corresponding}
\end{preprintauthorlist}
\preprintaffiliation{sjtu}{Shanghai Jiaotong University\quad}
\preprintaffiliation{ustc}{University of Science and Technology of China}
\expandafter\gdef\csname preprintcorrespondingauthor@text\endcsname{Zhouhan Lin}

\preprintkeywords{looped transformers, iterative computation, depth growing, pretraining efficiency}

\vskip 0.3in
]

\printAffiliationsAndNotice{}
\suppressfloats[t]

\begin{abstract}
Looped language models can strengthen reasoning through repeated computation with shared weights, increasing execution depth without adding independent parameters.
However, each additional pass increases training floating-point operations (FLOPs), making these gains more expensive to obtain.
To address this cost, we propose \textbf{UpLoop}, a pretraining schedule that uses upcycling to train looped Transformers while exactly matching a standard dense baseline in final parameter count, total training tokens, and training FLOPs.
Specifically, UpLoop begins with a shallow Transformer, progressively grows its depth by copying learned layers, and uses the computation saved early to support repeated execution of a shared middle block later in training.
Experiments across model scales show improvements in language modeling and downstream performance, with the average score across 22 downstream tasks increasing from 22.99 to 27.52 at 1.3B parameters.
These gains also extend to mathematical reasoning after supervised fine-tuning.
\end{abstract}

\section{Introduction}
\label{sec:intro}

\begin{figure}[t]
\centering
\includegraphics[width=\columnwidth]{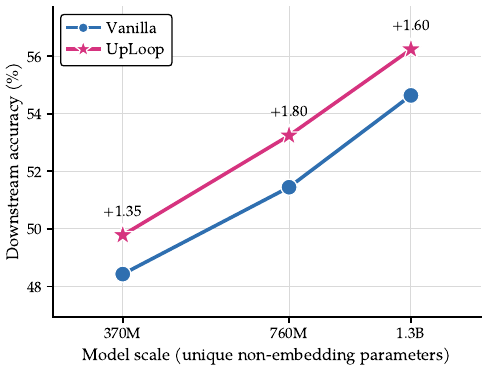}
\caption{Zero-shot accuracy of UpLoop and Vanilla across three model scales
from 370M to 1.3B parameters. Both methods are trained at 100 tokens per
parameter with equal training tokens and FLOPs at each scale. See
Appendix~\ref{app:downstream-eval} for evaluation details.}
\label{fig:scale}
\end{figure}

\begin{figure*}[t]
\centering
\includegraphics[width=\textwidth]{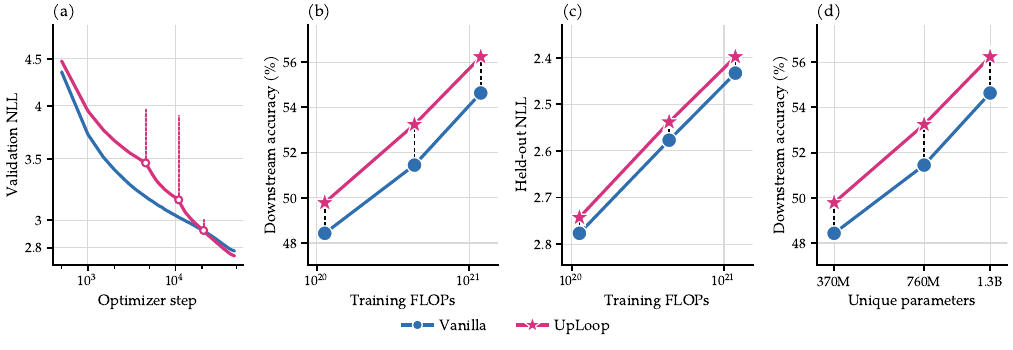}
\caption{Training dynamics and performance of UpLoop and Vanilla with equal
training tokens and FLOPs at each scale. \textbf{(a)} Validation NLL during
370M pre-training. \textbf{(b, c)} Downstream accuracy and held-out NLL versus
training FLOPs. \textbf{(d)} Downstream accuracy versus unique non-embedding
parameters. See Appendix~\ref{app:downstream-eval} for evaluation details.}
\label{fig:tradeoff}
\end{figure*}

Large language models (LLMs) have demonstrated strong capabilities in mathematical reasoning \citep{muennighoff2025s1,xin2025deepseek} and agentic coding \citep{wang2025openhands,yang2024swe,yang2026swe}.
These capabilities have advanced substantially with increases in model size and training data \citep{hoffmann2022training}.
Alongside scaling parameters and data, looped language models introduce an additional axis for scaling through iterative computation \citep{geiping2025scaling,zhu2025scaling,alabdulmohsin2026recursive,bae2025relaxed,asymloop}.
Specifically, they repeatedly apply a shared Transformer block to increase execution depth at a fixed number of independent parameters, allowing more rounds of computation that can strengthen reasoning \citep{saunshi2025reasoning}.

However, the benefits of repeated computation come with increased training FLOPs.
Each additional pass through the shared block requires extra computation, making a looped language model more expensive to train than its dense counterpart on the same number of tokens.
This raises a practical question: \textbf{Can we train a looped language model that outperforms a standard dense model with the same model parameters, training tokens, and training FLOPs?}

To accommodate the additional computation required by looping within the dense baseline's training budget, we introduce a pretraining schedule that reallocates computation across training stages.
A fixed total compute budget constrains the accumulated training cost, while allowing different stages to use different execution depths.
We allocate less computation to early tokens and progressively increase depth, carrying learned parameters forward across stages \citep{shen2022staged,du2024stacking,yao2023masked,saunshi2024inductive}.
The computation saved early then supports repeated execution in the final stage, allowing us to train a looped model within the dense baseline's original token and training FLOP budgets.

We instantiate this strategy as \textbf{UpLoop}, a pretraining schedule that combines progressive depth growth with shared-weight looping.
Specifically, UpLoop first trains a shallow Transformer and progressively copies its learned layers to increase depth.
Once the model reaches its target depth, it repeatedly executes a middle block with shared weights, gaining additional execution depth without adding model parameters.

Across three model scales, UpLoop improves held-out language modeling and average downstream performance over the dense baseline under equal training token and FLOP budgets.
Specifically, the average score across 22 downstream tasks increases from 22.99 to 27.52 at 1.3B parameters.
These gains also extend to downstream mathematical reasoning after supervised fine-tuning.
Further analyses show more aligned gradients across shared calls, functional differentiation among copied layers, and improved contextual retrieval.

\begin{figure*}[t]
\centering
\includegraphics{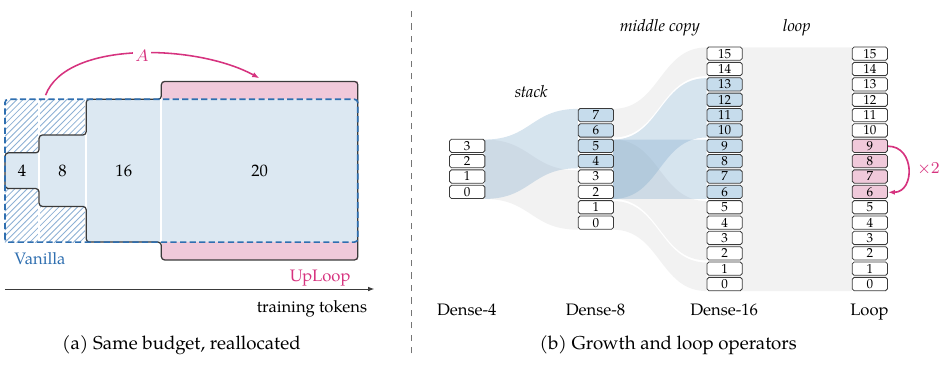}
\caption{Overview of UpLoop. \textbf{(a)} \textcolor[HTML]{2F6FB0}{Ribbon} thickness denotes the execution
depth over training tokens. \textcolor[HTML]{D6337F}{UpLoop} runs shallower than \textcolor[HTML]{2F6FB0}{Vanilla} early in training
and loops a middle block late, so the compute saved early (\textcolor[HTML]{D6337F}{$A$}) equals the extra
compute spent on looping. \textbf{(b)} Each stage is initialised by copying layers
of the previous stage; \textcolor[HTML]{595959}{grey bands} keep layers and \textcolor[HTML]{3F6F96}{blue bands} copy them.}
\label{fig:method}
\end{figure*}

\section{Related Work}
\label{sec:related}

\paragraph{Looped and recursive language models.}
Reusing a block along the depth axis goes back to Universal Transformers
\citep{dehghani2019universal} and has recently been revived as a way to buy
iterative computation at a fixed parameter count
\citep{yang2023looped,saunshi2025reasoning,geiping2025scaling,zhu2025scaling},
sometimes with the shared block relaxed by per-pass adapters or experts
\citep{csordas2024moeut,bae2025relaxed,bae2025mixture},
with the reported gains concentrated in reasoning
\citep{saunshi2025reasoning,kapl2026growing}. Because a looped model executes
more layers than the dense model it is compared against, those comparisons hold
parameters, and sometimes tokens, fixed and pay for the extra execution depth
with extra compute at both training and inference time. Concurrent scaling-law
work makes the dependence explicit: at matched parameters and data, looping
lowers loss \citep{prairie2026parcae}, but at matched \emph{computational depth}
more recurrence is worse once model size and data are jointly optimised
\citep{schwethelm2026much}, and compute-matched gains have so far been reported
for sparse looped mixtures of experts \citep{wang2026smelt}. Our budget is
stricter than any of these: both the tokens and the training FLOPs are the dense
baseline's, to the batch.

\paragraph{Depth growth and upcycling.}
A parallel line reuses a trained model to initialise a larger one by stacking or
duplicating layers
\citep{chen2015net2net,gong2019efficient,shen2022staged,wang2023learning,yao2023masked,kim2023solar,du2024stacking},
or by turning a dense checkpoint into a sparse one
\citep{komatsuzaki2023sparse}. The original motivation is saving training
FLOPs---\citet{du2024stacking} find depthwise stacking the most efficient
operator, and the grown models of \citet{kapl2025depth} reach the baseline's
final depth with roughly $77\%$ of its training compute---and
\citet{saunshi2024inductive} then showed that duplicating in the middle of the
stack carries an inductive bias toward reasoning, a view
\citet{kapl2026growing} extend into one picture of growing and looping as
implicit versus explicit repetition. Closest to the control we need,
\citet{chen2026growth} grow a looped core with its weights either tied or
untied at matched FLOPs and report that both improve the compute-optimal
scaling exponent, untied ahead of tied by a constant factor; the choice of
\emph{which} block to duplicate also matters \citep{karp2024landscape}.
Grown and upcycled models keep the duplicated weights independent and are dense
at inference. Growth can stop at the baseline's final depth, saving compute at
fixed tokens, or use the saved compute to train on more tokens at fixed FLOPs.
UpLoop spends the dense baseline's budget
\emph{exactly} and still ends up looped, which turns the comparison against that
baseline into an equal-budget one.

\section{Preliminary}
\label{sec:preliminary}

We first formulate the problem of training looped language models under fixed token and training FLOP budgets (Section~\ref{sec:task-formulation}).
We then review depth growth and looped Transformers (Section~\ref{sec:depth-growth-and-looping}).

\subsection{Task Formulation}
\label{sec:task-formulation}

A standard Transformer applies $L$ layers sequentially:
\begin{equation}
\label{eq:dense-forward}
\mathbf h^{(i)}=F_{\theta_i}\!\left(\mathbf h^{(i-1)}\right),
\quad i=1,\ldots,L,
\end{equation}
where $F_{\theta_i}$ denotes the $i$-th Transformer layer.
A looped Transformer instead repeatedly applies a shared block $B_\phi$:
\begin{equation}
\label{eq:looped-forward}
\mathbf z^{(k)}=B_\phi\!\left(\mathbf z^{(k-1)}\right),
\quad k=1,\ldots,r,
\end{equation}
where $\mathbf z^{(0)}$ is the block input and the parameters $\phi$ are shared across all $r$ calls.
Our goal is to train a looped language model with the same total training tokens $D_0$ and training FLOPs $C_0$ as a standard dense model.

\subsection{Depth Growth and Looped Transformers}
\label{sec:depth-growth-and-looping}

Depth growth expands a shallow Transformer by copying existing layers during training.
To grow from $d$ to $d'>d$ layers, each destination layer is initialized as:
\begin{equation}
\label{eq:depth-growth-copy}
\theta_i' \leftarrow \operatorname{copy}\!\left(\theta_{\pi(i)}\right),
\quad i=1,\ldots,d',
\end{equation}
where $\pi(i)$ specifies the source layer.
The copied layers are optimized independently after initialization.
Using fewer layers early in training saves FLOPs relative to training the full-depth model throughout.

Looping provides a way to scale computation in language models by repeatedly applying shared weights.
This additional computation can improve model performance while keeping the number of independent parameters fixed.
However, it also increases training FLOPs per token, making looping more expensive at the same token budget.
This trade-off motivates using the computation saved during early shallow training to support later looping.

\section{Methodology} \label{sec:method}

We present UpLoop, a four-stage pretraining schedule that combines progressive depth growth with middle-block looping, as shown in Figure~\ref{fig:method}.
Specifically, UpLoop progressively expands a shallow Transformer through layer copying and trains the model at each depth before enabling shared-weight looping in the final stage.
Through progressive depth growth and the allocation of training tokens across stages, UpLoop accommodates the additional computation required by looping while exactly matching the dense baseline in final parameter count, total training tokens, and analytical training FLOPs.

UpLoop follows a four-stage training schedule with execution depths:
\begin{equation}\label{eq:stage-depth-schedule}
\frac{L}{4}\;\rightarrow\;\frac{L}{2}\;\rightarrow\;L\;\rightarrow\;L+(r-1)m,
\end{equation}
where $L$ is the dense baseline's depth, $m$ is the shared middle block's depth, and $r$ is its repetition count in the final stage.
We first describe progressive depth growth (Section~\ref{sec:progressive-depth-growth}) and middle-block looping (Section~\ref{sec:middle-block-looping}), then describe the training schedule (Section~\ref{sec:equal-budget-stage-allocation}).

\subsection{Progressive Depth Growth} \label{sec:progressive-depth-growth}
To reuse learned parameters during depth growth, UpLoop initializes each deeper model by copying layers from the preceding stage.
Specifically, the first growth transition stacks the $L/4$ Transformer layers twice to initialize an $L/2$-layer model.
Each copy inherits both the layer weights and their corresponding Adam states, and the duplicated layers are subsequently optimized independently.

To grow from $L/2$ to $L$ layers, UpLoop preserves the head and tail and expands the middle block.
Specifically, we partition the preceding model into three parameter blocks $(\Theta_{\mathrm H},\Theta_{\mathrm B},\Theta_{\mathrm T})$, with depths $(L/8,L/4,L/8)$, and initialize the next stage as:
\begin{equation}\label{eq:middle-block-growth}
\Theta_L \leftarrow \bigl(\Theta_{\mathrm H},\Theta_{\mathrm B}^{[1]},\Theta_{\mathrm B}^{[2]},\Theta_{\mathrm B}^{[3]},\Theta_{\mathrm T}\bigr),
\end{equation}
where each $\Theta_{\mathrm B}^{[j]}$ is a separate copy of $\Theta_{\mathrm B}$.
As in the first transition, both layer weights and Adam states are copied, and all layers are subsequently optimized independently.

\subsection{Middle-Block Looping} \label{sec:middle-block-looping}
To increase execution depth without adding parameters, UpLoop enables middle-block looping after the full-depth dense stage.
Specifically, we select $m$ consecutive middle layers as a shared block $B_\phi$, with the remaining layers forming a head $F_{\mathrm H}$ and a tail $F_{\mathrm T}$.
The forward pass becomes:
\begin{equation}\label{eq:uploop-forward}
\mathbf h_{\mathrm{out}}=F_{\mathrm T}\!\left(B_\phi^{\circ r}\!\left(F_{\mathrm H}(\mathbf h_{\mathrm{in}})\right)\right),
\end{equation}
where $B_\phi^{\circ r}$ denotes $r$ successive calls with shared weights, while the head and tail are each executed once.
This transition preserves all parameters and optimizer states while increasing execution depth to $L+(r-1)m$.

\subsection{Compute-Matched Training Schedule} \label{sec:equal-budget-stage-allocation}
We compute UpLoop's total training cost by summing the contributions of its four stages.
Specifically, for stage $j$ with $D_j$ training tokens and execution depth $d_j$, the total token count $D$ and training FLOPs $C$ are:
\begin{equation}\label{eq:training-cost-accounting}
D=\sum_{j=1}^{4}D_j,
\qquad
C=\sum_{j=1}^{4}D_j f(d_j),
\end{equation}
where $f(d)$ denotes the analytical training FLOPs per token at execution depth $d$.
UpLoop's stage allocation yields $D=D_0$ and $C=C_0$, preserving the dense baseline's total token and training FLOP budgets while enabling greater execution depth in the final stage.
The complete stage schedules and FLOP accounting are provided in Appendix~\ref{app:budgets}.

UpLoop uses a common schedule across model scales.
Specifically, the transitions to the second, third, and final stages occur after approximately $9.7\%$, $23.0\%$, and $44.3\%$ of the total training tokens, respectively.
Training proceeds on a continuous data stream, and the learning-rate schedule follows cumulative training tokens without restarting at stage boundaries.

\begin{table*}[t]
\caption{Zero-shot accuracy (\%) of UpLoop and Vanilla from 370M to 1.3B
parameters with equal training tokens and FLOPs at each scale. See
Appendix~\ref{app:downstream-eval} for evaluation details.}
\label{tab:commonsense}
\vskip 0.1in
\centering
\begin{small}
\setlength{\tabcolsep}{3.5pt}
\begin{tabular}{llrrrrrrrrr}
\toprule
Scale & Model & OBQA & WinoG. & ARC-c & ARC-e & HellaSwag & SIQA & PIQA & BoolQ & Average \\
\midrule
370M & Vanilla & 20.80$^{\dagger}$ & 52.01$^{\dagger}$ & 28.84 & 62.75 & 35.02 & 38.69 & 67.90 & 57.37 & 48.43 \\
 & UpLoop & 23.80$^{\dagger}$ & 53.28$^{\dagger}$ & \textbf{29.01} & \textbf{63.22} & \textbf{36.02} & \textbf{40.17} & \textbf{68.66} & \textbf{61.62} & \textbf{49.78} \\
\midrule
760M & Vanilla & 26.80$^{\dagger}$ & 55.01 & 32.17 & 68.06 & 39.67 & 39.66 & 70.78 & 58.35 & 51.96 \\
 & UpLoop & 26.20$^{\dagger}$ & \textbf{55.17} & \textbf{32.85} & \textbf{69.57} & \textbf{41.58} & \textbf{42.07} & \textbf{70.89} & \textbf{62.51} & \textbf{53.52} \\
\midrule
1.3B & Vanilla & 28.40$^{\dagger}$ & 56.51 & \textbf{37.54} & \textbf{73.36} & 43.21 & 40.84 & 73.12 & 59.76 & 54.91 \\
 & UpLoop & 30.40$^{\dagger}$ & \textbf{57.85} & 37.20 & 73.23 & \textbf{44.84} & \textbf{42.37} & \textbf{73.72} & \textbf{66.06} & \textbf{56.47} \\
\bottomrule
\end{tabular}
\end{small}
\vskip -0.1in
\end{table*}

\section{Experiments}
\label{sec:exp}

In this section, we begin by presenting the experimental setup
(\cref{subsec:setup}), followed by a comparison between UpLoop and the dense
baseline under equal training tokens and FLOPs (\cref{subsec:equal}).

\subsection{Experimental Setup}
\label{subsec:setup}

\textbf{Models and Datasets.} We evaluate UpLoop through language modeling
and downstream tasks across three model scales.
\textbf{(i) Pre-training.} We pre-train decoder-only Transformers based on the
OLMo~2 Ladder architecture \citep{olmo2025olmo2} on FineWeb-Edu
\citep{penedo2024fineweb}, with 370M, 760M and 1.3B non-embedding
parameters. All models are trained at 100 tokens per parameter, five times
the Chinchilla token budget. At each scale,
UpLoop and Vanilla use the same training tokens and total training FLOPs.
Detailed model configurations and training budgets are provided in
Appendix~\ref{app:budgets}.
\textbf{(ii) Downstream evaluation.} We evaluate the pretrained checkpoints
using two benchmark suites. First, we evaluate all three model scales on eight
zero-shot tasks: PIQA~\citep{bisk2020piqa}, SIQA~\citep{sap2019socialiqa},
HellaSwag~\citep{zellers2019hellaswag}, WinoGrande~\citep{sakaguchi2021winogrande},
ARC-c~\citep{clark2018think}, ARC-e~\citep{clark2018think},
OBQA~\citep{mihaylov2018can} and
BoolQ~\citep{boolq}, using the standard prompts in lm-evaluation-harness
\citep{gao2024lmeval}. Second, we evaluate the 370M, 760M and 1.3B models on
22 downstream tasks covering closed-book and open-book question answering,
language understanding, math word problems and reasoning primitives,
following \citet{kapl2026growing}. The mathematics and reasoning primitive
tasks use five-shot evaluation. Detailed evaluation protocols and aggregation
rules are provided in Appendix~\ref{app:downstream-eval}.
\textbf{(iii) Language modeling.} We evaluate language modeling performance
using negative log-likelihood (NLL) on a fixed held-out set of 2,048
FineWeb-Edu~\citep{penedo2024fineweb} sequences, each containing 4,096 tokens, shared across all models.

\textbf{Baselines.} We compare UpLoop with the following baselines:
(i) \textsc{Vanilla}, a standard dense Transformer following OLMo~2;
(ii) \textbf{Loop from scratch}, a model trained with the looped architecture
throughout training, starting from random initialization;
(iii) \textbf{Loop with dense initialization}, which first trains a full-depth
dense model and then continues training with looping;
(iv) \textbf{Dense-20}, which uses independent layers to match UpLoop's final
execution depth, with either matched training tokens or matched training FLOPs; and
(v) \textbf{LIDAS}, which progressively grows a dense model without
looping and uses the saved compute to train on additional tokens.
Detailed budget configurations are provided in Appendix~\ref{app:training-budget-comparisons}.

\subsection{LLM Pre-training}
\label{subsec:pretraining}

\subsubsection{Results across Model Scales}
\label{subsec:equal}

\textbf{Language Modeling.}
Figure~\ref{fig:tradeoff} and Appendix Table~\ref{tab:nll} compare UpLoop with
Vanilla across three model scales, from 370M to 1.3B, under equal training token
and FLOP budgets. UpLoop achieves lower NLL at all three scales. At 370M, it
reduces NLL from 2.78 to 2.74, while at 1.3B, NLL decreases from 2.43 to
2.40. These results demonstrate that UpLoop consistently improves language
modeling across all model sizes under equal training budgets.

\textbf{Zero-shot Commonsense Reasoning.}
Table~\ref{tab:commonsense} reports zero-shot performance on eight benchmarks
under equal training token and FLOP budgets. At 370M, 760M and 1.3B, UpLoop
improves average accuracy from $48.43\%$ to $49.78\%$, from $51.96\%$ to
$53.52\%$, and from $54.91\%$ to $56.47\%$, respectively. It outperforms
Vanilla on all six eligible tasks at 370M, all seven at 760M, and five of seven
at 1.3B. These results show that the pretraining gains consistently translate
into improved zero-shot performance across all evaluated model scales.

\textbf{Reasoning and Knowledge Evaluation.}
\begin{table*}[t]
\caption{Downstream performance of UpLoop and Vanilla on knowledge, language,
mathematics and reasoning tasks with equal training tokens and FLOPs at each
scale. See Appendix~\ref{app:downstream-eval} for evaluation details.}
\label{tab:main}
\vskip 0.1in
\centering
\begin{small}
\setlength{\tabcolsep}{7pt}
\begin{tabular}{lrrrrrr}
\toprule
& \multicolumn{2}{c}{370M} & \multicolumn{2}{c}{760M} & \multicolumn{2}{c}{1.3B} \\
\cmidrule(lr){2-3}\cmidrule(lr){4-5}\cmidrule(lr){6-7}
Task & Vanilla & UpLoop & Vanilla & UpLoop & Vanilla & UpLoop \\
\midrule
\multicolumn{7}{l}{\textit{Closed-book question answering}} \\
TriviaQA & 13.36 & \textbf{13.72} & 20.75 & \textbf{20.99} & \textbf{22.24} & 20.79 \\
TyDiQA-NoContext & 8.66 & \textbf{8.81} & \textbf{13.36} & 12.39 & 12.71 & \textbf{12.84} \\
NaturalQuestions & 5.10 & \textbf{5.53} & 7.93 & \textbf{8.69} & \textbf{9.43} & 9.04 \\
WebQuestions & 10.85 & \textbf{11.17} & 14.67 & \textbf{16.76} & \textbf{17.13} & 15.00 \\
\midrule
\multicolumn{7}{l}{\textit{Open-book question answering}} \\
TyDiQA-GoldP & 15.17 & \textbf{16.46} & 18.22 & \textbf{22.48} & \textbf{22.44} & 22.28 \\
SQuADv2 & 6.35$^{\dagger}$ & 7.01$^{\dagger}$ & 9.49 & \textbf{10.70} & 11.09 & \textbf{11.52} \\
DROP & 7.09 & \textbf{7.60} & 8.70 & \textbf{10.74} & 10.12 & \textbf{12.18} \\
QuAC & 23.04 & \textbf{24.88} & 27.46 & \textbf{29.32} & 30.59 & \textbf{32.72} \\
CoQA & 27.51 & \textbf{29.12} & 34.20 & \textbf{38.93} & 38.85 & \textbf{44.38} \\
\midrule
\multicolumn{7}{l}{\textit{Language understanding}} \\
LAMBADA & 34.02 & \textbf{34.58} & 40.44 & \textbf{42.89} & 45.57 & \textbf{49.49} \\
HellaSwag & 42.60 & \textbf{44.93} & 50.34 & \textbf{53.46} & 56.40 & \textbf{59.00} \\
\midrule
\multicolumn{7}{l}{\textit{Math word problems}} \\
SVAMP & 1.61$^{\dagger}$ & 2.01$^{\dagger}$ & 3.52 & \textbf{3.92} & 5.03 & \textbf{5.93} \\
ASDiv & 1.97 & \textbf{2.52} & \textbf{5.73} & 5.13 & 6.18 & \textbf{9.94} \\
MAWPS AddSub & 1.39$^{\dagger}$ & 1.39$^{\dagger}$ & \textbf{3.88} & 2.77 & 3.60 & \textbf{5.82} \\
MAWPS MultiArith & 2.96$^{\dagger}$ & 4.52$^{\dagger}$ & 4.52 & \textbf{4.70} & 4.70 & \textbf{5.57} \\
MAWPS SingleOp & 1.14$^{\dagger}$ & 2.66$^{\dagger}$ & 4.37 & \textbf{7.79} & 8.17 & \textbf{10.84} \\
MAWPS SingleEq & 0.83$^{\dagger}$ & 0.83$^{\dagger}$ & 2.49$^{\dagger}$ & 3.88$^{\dagger}$ & 4.43 & \textbf{7.20} \\
\midrule
\multicolumn{7}{l}{\textit{Reasoning primitives}} \\
Copying (random words) & 11.60$^{\dagger}$ & 12.90$^{\dagger}$ & \textbf{15.50} & 13.90 & 18.60 & \textbf{21.60} \\
Copying (real words) & 11.30$^{\dagger}$ & 17.00$^{\dagger}$ & 15.50 & \textbf{17.40} & 23.80 & \textbf{27.50} \\
VarAssign-basic & 20.30$^{\dagger}$ & 23.90$^{\dagger}$ & \textbf{31.50} & 28.00 & 30.00 & \textbf{39.60} \\
VarAssign-math & 28.10 & \textbf{34.40} & 59.90 & \textbf{82.80} & 58.10 & \textbf{91.10} \\
VarAssign-code & 28.90 & \textbf{38.00} & 59.10 & \textbf{87.50} & 66.60 & \textbf{91.20} \\
\midrule
\textbf{Average} & 18.95 & \textbf{20.90} & 21.38 & \textbf{24.82} & 22.99 & \textbf{27.52} \\
Valid tasks & \multicolumn{2}{c}{13/22} & \multicolumn{2}{c}{21/22} & \multicolumn{2}{c}{22/22} \\
\bottomrule
\end{tabular}
\end{small}
\vskip -0.1in
\end{table*}

Table~\ref{tab:main} compares UpLoop with Vanilla on 22 downstream tasks
covering question answering, language understanding, mathematics and reasoning
under equal training token and FLOP budgets. At 370M, 760M and 1.3B, UpLoop
improves the average score from 18.95 to 20.90, from 21.38 to 24.82, and from
22.99 to 27.52, respectively. It outperforms Vanilla on 13 of 13, 16 of 21,
and 18 of 22 eligible tasks at the three scales. These results demonstrate
that UpLoop improves overall downstream performance across all evaluated
model scales.

The largest gains in Table~\ref{tab:main} occur on reasoning primitives,
particularly variable assignment. At 1.3B, UpLoop improves accuracy from
$58.10\%$ to $91.10\%$ on VarAssign-math and from $66.60\%$ to $91.20\%$ on
VarAssign-code. Open-book question answering also benefits, with CoQA F1
increasing from 38.85 to 44.38 at the same scale. These results highlight
UpLoop's advantage in strengthening reasoning and context-based question
answering under equal training token and FLOP budgets.

UpLoop also improves performance on math word problems. At 1.3B, it outperforms
Vanilla on all six eligible benchmarks, increasing average accuracy from
$5.35\%$ to $7.55\%$. For example, accuracy improves from $6.18\%$ to $9.94\%$
on ASDiv and from $4.43\%$ to $7.20\%$ on MAWPS SingleEq. These results show
that UpLoop's benefits extend to mathematical reasoning under equal training
budgets.

\subsubsection{Comparison with Alternative Training Strategies}
\label{sec:training-strategy-comparison}
\label{subsubsec:training-strategies}

\begin{table}[t]
\caption{Comparison with alternative training strategies at 370M. Training
tokens and FLOPs are normalized to UpLoop.}
\label{tab:training-strategies}
\vskip 0.1in
\centering
\begin{small}
\setlength{\tabcolsep}{3pt}
\renewcommand{\arraystretch}{1.1}
\begin{tabular}{lrrrr}
\toprule
Model & Params. & Tokens & FLOPs & NLL $\downarrow$ \\
\midrule
Vanilla & 371M & $1.00\times$ & $1.00\times$ & 2.78 \\
Loop from scratch & 371M & $1.00\times$ & $1.20\times$ & 2.76 \\
\shortstack[l]{Loop with dense\\initialization} & 371M & $1.00\times$ & $1.11\times$ & 2.76 \\
\shortstack[l]{Dense-20\\(Iso-FLOPs)} & 438M & $0.83\times$ & $1.00\times$ & 2.76 \\
\shortstack[l]{Dense-20\\(Iso-Tokens)} & 438M & $1.00\times$ & $1.20\times$ & \textbf{2.74} \\
\shortstack[l]{LIDAS (Iso-FLOPs)\\\citep{kapl2025depth}} & 371M & $1.20\times$ & $1.00\times$ & 2.75 \\
\midrule
\textbf{UpLoop} & 371M & $1.00\times$ & $1.00\times$ & \textbf{2.74} \\
\bottomrule
\end{tabular}
\end{small}
\vskip -0.1in
\end{table}

To assess the contributions of staged growth and looping, we compare UpLoop
with looping, dense and depth-growth baselines at 370M. All methods use the
same model width and optimization recipe; Table~\ref{tab:training-strategies}
summarizes their parameter counts, training token budgets, training FLOPs and
NLL.

\textbf{Looping Baselines.}
Table~\ref{tab:training-strategies} compares UpLoop with Loop from scratch and
Loop with dense initialization at the same parameter count and training token
budget. All three models share the same final looped architecture. The two
baselines require $1.20\times$ and $1.11\times$ the training FLOPs of UpLoop,
respectively, and both achieve an NLL of 2.76, compared with 2.74 for UpLoop.
These results show that UpLoop's staged schedule improves language modeling
with fewer training FLOPs.

\textbf{Dense and Growth Baselines.}
We compare UpLoop with two Dense-20 settings, both using 20 independent layers
to match its final execution depth. \textbf{Dense-20 (Iso-FLOPs)} matches
UpLoop's training FLOP budget, uses $0.83\times$ the training tokens, and
achieves an NLL of 2.76, compared with 2.74 for UpLoop.
\textbf{Dense-20 (Iso-Tokens)} matches the training token budget and achieves
an NLL of 2.74, requiring $1.20\times$ the training FLOPs. Both Dense-20
settings use 438M non-embedding parameters, compared with 371M for UpLoop.
These results highlight UpLoop's advantage in achieving lower NLL with fewer
parameters under a fixed training FLOP budget.

We further compare UpLoop with \textbf{LIDAS (Iso-FLOPs)}, which progressively
grows a dense model without looping. With training FLOPs and parameter count
matched, LIDAS uses $1.20\times$ the training tokens and achieves an NLL of
2.75, compared with 2.74 for UpLoop. This result shows that UpLoop achieves
lower NLL with fewer training tokens than the growth-only baseline.

\subsection{Supervised Fine-tuning on GSM8K}
\label{subsec:gsm8k-sft}

We evaluate whether UpLoop's pretraining gains carry over to supervised
fine-tuning on GSM8K. We fine-tune the 1.3B Vanilla and UpLoop models for
three epochs using the same training data and optimization hyperparameters,
retaining UpLoop's shared-weight architecture during full-parameter
fine-tuning. As shown in Table~\ref{tab:gsm8k-sft}, UpLoop improves
exact-match accuracy from $8.72\%$ to $12.05\%$ under zero-shot evaluation
and from $5.46\%$ to $8.11\%$ under five-shot evaluation. These results
demonstrate that UpLoop's benefits extend to task-specific fine-tuning.
Detailed training and evaluation settings are provided in
Appendix~\ref{app:gsm8k-sft}.

\begin{table}[ht]
\caption{GSM8K exact-match accuracy after supervised fine-tuning at 1.3B
parameters. Experimental details are provided in
Appendix~\ref{app:gsm8k-sft}.}
\label{tab:gsm8k-sft}
\vskip 0.1in
\centering
\begin{small}
\begin{tabular}{lccc}
\toprule
Model & Params. & Zero-shot (\%) $\uparrow$ & Five-shot (\%) $\uparrow$ \\
\midrule
Vanilla & 1.3B & 8.72 & 5.46 \\
UpLoop  & 1.3B & \textbf{12.05} & \textbf{8.11} \\
\bottomrule
\end{tabular}
\end{small}
\vskip -0.1in
\end{table}

\subsection{Further Analysis}
\label{subsec:analysis}

To understand these performance gains, we further analyze the final pretrained models. We first compare UpLoop with the looping baselines in gradient alignment across calls and the propagation of state perturbations, then examine functional inheritance and differentiation after depth growth, and finally analyze contextual retrieval performance and the contribution of second-pass reading to answer formation. Experimental details are provided in Appendix~\ref{app:analysis-details}.

\begin{figure}[ht]
\centering
\includegraphics[width=\columnwidth]{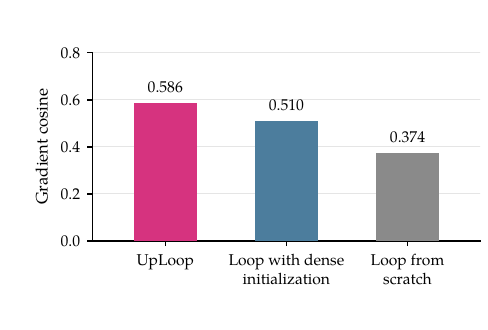}
\caption{\textbf{Gradient alignment across calls.} In the final 370M models, bars show mean cosine similarity between prediction-loss gradients from two calls to each shared layer, averaged over four shared layers and 12 windows.}
\label{fig:analysis-gradient}
\end{figure}

\textbf{More aligned gradients across calls.} To assess the compatibility of optimization demands across calls, we isolate the gradient contributions of the two calls to each shared layer at the final checkpoint and compute their cosine similarity. Figure~\ref{fig:analysis-gradient} shows that UpLoop achieves a mean gradient cosine of \textbf{0.586}, compared with \textbf{0.510} for Loop with dense initialization and \textbf{0.374} for Loop from scratch, indicating more aligned update directions for shared parameters across calls.

\textbf{More robust propagation of state deviations.} We inject a small state perturbation after the first call to the shared block and measure the distance between the perturbed and unperturbed trajectories. As shown in Table~\ref{tab:analysis-perturbation}, after the normal two calls to the shared block, UpLoop reduces this distance to \textbf{0.867 times} the initial perturbation distance, compared with \textbf{0.933} for Loop with dense initialization and \textbf{0.923} for Loop from scratch. When execution is extended to eight calls, all three models amplify the tested deviations, with UpLoop showing the least amplification.

\begin{table}[ht]
\caption{\textbf{Perturbation propagation.} In the 370M models, we inject a single perturbation after the first call to the shared block and report trajectory distance divided by the initial perturbation distance. Results average 512 windows after the normal two calls and an extended eight calls to the shared block.}
\label{tab:analysis-perturbation}
\vskip 0.1in
\centering
\begingroup
\small
\setlength{\tabcolsep}{4pt}
\begin{tabular}{lcc}
\toprule
Model & \begin{tabular}[c]{@{}c@{}}Normal\\2 calls $\downarrow$\end{tabular} & \begin{tabular}[c]{@{}c@{}}Extended\\8 calls $\downarrow$\end{tabular} \\
\midrule
UpLoop & \textbf{0.867} & \textbf{1.906} \\[3pt]
Loop with dense initialization & 0.933 & 2.965 \\[3pt]
Loop from scratch & 0.923 & 2.644 \\[3pt]
\bottomrule
\end{tabular}
\endgroup
\vskip -0.1in
\end{table}

\textbf{Functional inheritance and differentiation after depth growth.} UpLoop increases depth by copying existing layers, allowing the added layers to continue training from previously learned parameters. In 370M UpLoop, distinct layers copied from the same source at the final growth transition have a mean weight cosine similarity of \textbf{0.383} at the end of training, down from 1 at initialization and substantially above \textbf{0.024} for the remaining layer pairs. Thus, the layers produced by depth growth develop distinct parameter configurations while retaining relationships inherited from their common source.

\begin{figure}[ht]
\centering
\includegraphics[width=\columnwidth]{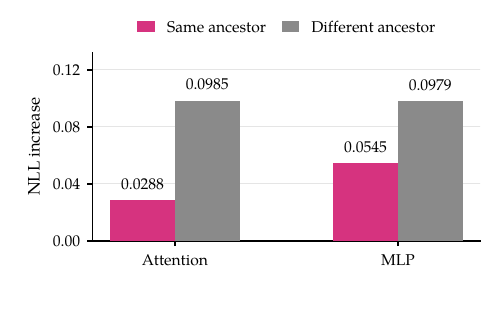}
\caption{\textbf{Layer substitution.} In 370M UpLoop, bars show mean $\Delta\mathrm{NLL}$ relative to native execution for substitutions between distinct layers with shared or different initial ancestry. Attention or MLP updates in the first call to the shared block are replaced using the same input and tokenwise norm matching. Results average 64 windows.}
\label{fig:analysis-lineage}
\end{figure}

To examine how ancestry relates to layer function, we trace each layer through the full growth history to its initial source and compare substitutions within and across ancestry groups. A replacement layer processes the target layer's input, and its attention or feed-forward (MLP) update replaces the original after tokenwise norm matching. Figure~\ref{fig:analysis-lineage} shows that substitutions between layers sharing an initial ancestor incur mean $\Delta\mathrm{NLL}$s of \textbf{0.0288} and \textbf{0.0545} for attention and MLP, respectively, compared with \textbf{0.0985} and \textbf{0.0979} for layers with different ancestors. The lower yet nonzero substitution costs among layers with shared ancestry indicate that these layers retain a common computational basis while developing distinct computations.

\begin{figure}[ht]
\centering
\includegraphics[width=\columnwidth]{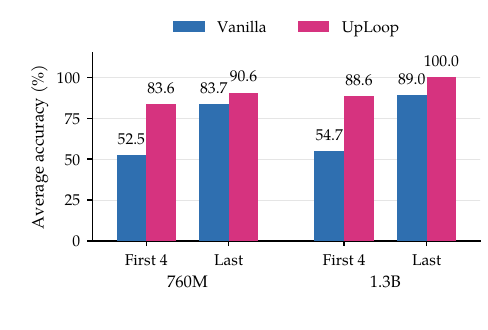}
\caption{\textbf{VarAssign accuracy by target position.} First 4 and Last denote targets among the first four assignments and in the last of five assignments, respectively. Results equally average Math and Code, with 1,000 examples per task.}
\label{fig:analysis-position}
\end{figure}

\textbf{Contextual retrieval and second-pass reading.} We further analyze VarAssign-math and VarAssign-code, which show substantial gains in Table~\ref{tab:main}, referring to them as Math and Code, respectively. Both tasks present five variable assignments and ask for the value of a queried variable. Figure~\ref{fig:analysis-position} reports their equally weighted mean accuracy by target assignment position. For targets among the first four assignments, UpLoop achieves \textbf{83.6\% and 88.6\%} at 760M and 1.3B, respectively, compared with \textbf{52.5\% and 54.7\%} for Vanilla; the corresponding gains for the final assignment are \textbf{6.9 and 11.0} percentage points. UpLoop thus narrows the accuracy gap between earlier and final assignments.

To examine the role of repeated execution, we block attention from the query portion to the target value in all four layers of the shared block during its second call, using an unrelated value as the control. Table~\ref{tab:analysis-reading} shows that average accuracy decreases by \textbf{8.6 and 14.5} percentage points relative to this control at 760M and 1.3B, respectively. These results show that the second call to the shared core supports answer formation through target-information reading; Appendix~\ref{app:reading-stages} compares the two core calls.

\begin{table}[ht]
\caption{\textbf{UpLoop accuracy (\%) under second-pass reading blocks.} Native denotes unmodified execution; Target and Control block query access to the target and an unrelated value, respectively, in all four layers of the shared block during its second call. Results equally average Math and Code, with 128 examples per task. Drop is Control minus Target, in percentage points.}
\label{tab:analysis-reading}
\vskip 0.1in
\centering
\begingroup
\fontsize{8}{9.5}\selectfont
\setlength{\tabcolsep}{4pt}
\begin{tabular}{lrrrr}
\toprule
Scale & Native & Control & Target & Drop (pp) \\
\midrule
760M & 84.8 & 86.3 & 77.7 & 8.6 \\[3pt]
1.3B & 89.1 & 89.5 & 75.0 & 14.5 \\[3pt]
\bottomrule
\end{tabular}
\endgroup
\vskip -0.1in
\end{table}

\section{Conclusion}\label{sec:conclusion}

We presented UpLoop, a pretraining schedule that trains looped Transformers within a standard dense model's parameter, token, and training FLOP budgets.
Specifically, UpLoop combined progressive depth growth with shared-weight looping, using computation saved during early shallow training to support repeated execution later.
Experiments across three model scales showed improvements in language modeling and average downstream performance, with benefits extending to mathematical reasoning after supervised fine-tuning.
Further analyses revealed more aligned gradients across shared calls, functional differentiation among copied layers, and improved contextual retrieval.
These findings supported reallocating computation across training stages as an effective approach to training looped language models without additional pretraining compute.
\label{main-text-end}

\newpage
\appendix
\raggedbottom
\makeatletter
\setlength{\@fptop}{0pt}
\setlength{\@fpsep}{14pt}
\setlength{\@fpbot}{0pt plus 1fil}
\makeatother
\setcounter{table}{0}
\renewcommand{\thetable}{A\arabic{table}}

\twocolumn[{
\begin{minipage}{\textwidth}
\section{Scales, Budgets and Costs}
\label{app:budgets}

\captionof{table}{\textbf{(Pre-training setup).} Shapes, budgets and hardware at the
three scales. $N$ counts every parameter except the input embedding and includes
the untied LM head; the feed-forward width is $4h$, the sequence length is
$4096$ at every scale, and a batch of $B$ sequences is $4096B$ supervised
tokens. $D_0$ is the token budget at $100$ tokens per parameter rounded up to a
whole global batch, and the peak learning rate is the OLMo~2 Ladder value for
that width, used unchanged by every method in this paper. Warmup is given in
updates and spans as many tokens as the model has parameters. MBS/GA is the
micro-batch size and the number of gradient-accumulation steps, listed for the
dense baseline and for UpLoop; the global batch, and therefore the optimizer
step, is identical for both. $^{\dagger}$Only the 1.3B pair shards optimizer
state (ZeRO-1).}
\label{tab:hparams}
\label{tab:budget}
\vskip 0.1in
\centering
\begin{small}
\setlength{\tabcolsep}{4.5pt}
\begin{tabular}{lrrrrrrrrrc}
\toprule
Scale & $N$ & $h$ & $L$ & Batch (seq.) & Updates & $D_0$ & Peak LR & Warmup
      & Dies & MBS/GA (V, UpLoop) \\
\midrule
370M & 371,262,464  & 1024 & 16 & 192 & 47,209 & 37.13\,B  & $7.8\times10^{-4}$ & 472 & 64 & 3/1,\ 1/3 \\
760M & 758,224,896  & 1536 & 16 & 320 & 57,848 & 75.82\,B  & $6.1\times10^{-4}$ & 578 & 64 & 1/5,\ 1/5 \\
1.3B & 1,279,395,840 & 2048 & 16 & 384 & 81,342 & 127.94\,B & $5.2\times10^{-4}$ & 813 & 64 & 2/3,\ 2/3$^{\dagger}$ \\
\bottomrule
\end{tabular}
\end{small}
\vskip -0.1in

\end{minipage}
\vspace{0.1in}
}]

\textbf{Training budget accounting.}
At a fixed model width and sequence length, analytical training FLOPs per token are affine in execution depth, $f(d)=\kappa d+\beta$, where $\kappa$ is the per-layer cost and $\beta$ is the depth-independent cost.
Since $\sum_{j=1}^{4}D_j=D_0$, the difference from the dense baseline is:
\begin{equation}\label{eq:appendix-budget-difference}
C-C_0=\kappa\left(\sum_{j=1}^{4}D_jd_j-LD_0\right).
\end{equation}
Substituting the four execution depths from Equation~\ref{eq:stage-depth-schedule}, matching training FLOPs therefore requires:
\begin{equation}\label{eq:appendix-budget-balance}
\frac{3L}{4}D_1+\frac{L}{2}D_2=(r-1)mD_4.
\end{equation}
The shallow-stage savings on the left exactly offset the looping overhead on the right.
Our experiments use $L=16$, $m=4$, and $r=2$, giving $12D_1+8D_2=4D_4$.
The four stage lengths in optimizer updates are $(4560,\allowbreak 6305,\allowbreak 10054,\allowbreak 26290)$ at 370M, $(5588,\allowbreak 7726,\allowbreak 12318,\allowbreak 32216)$ at 760M, and $(7857,\allowbreak 10863,\allowbreak 17325,\allowbreak 45297)$ at 1.3B.
With the global batch fixed within each scale, all three integer schedules satisfy both budget equalities exactly.

\textbf{Hyperparameters.} All pretraining experiments use AdamW with $\beta = (0.9, 0.95)$,
$\epsilon = 10^{-8}$ and weight decay $0.1$ on every parameter except the input
embedding, gradient clipping at $1.0$, and an auxiliary $z$-loss of $10^{-5}$.
The learning rate warms up linearly over as many tokens as the model has
non-embedding parameters and then decays by a cosine to $10\%$ of its peak; the
schedule runs on a token clock that is identical for both arms, so a stage
switch changes the architecture and never the learning rate. Peak learning
rates are the Ladder values for each width (\cref{tab:hparams}) and are not
retuned for UpLoop, which is deliberate: the claim is about reallocating a
budget, so the baseline keeps the configuration it was tuned for and UpLoop
inherits it unchanged.
Compute is BF16 with FP32 master weights and Adam states, and activation recomputation is off.
We periodically evaluate held-out NLL during pretraining to monitor training dynamics.
Downstream evaluations use the final checkpoint of each run.
Every run uses seed $42$ unless stated otherwise.

\begin{table}[!t]
\caption{Held-out NLL at the end of training, at exactly equal tokens and
equal training FLOPs; lower is better and \textbf{bold} marks the better arm at
each scale. Values are the native training-time final validation scores;
the FP32 conversion scores plotted in \cref{fig:tradeoff}(c) differ by less
than $0.0002$.}
\label{tab:nll}
\vskip 0.1in
\centering
\begin{small}
\begin{tabular}{lccc}
\toprule
Scale & Vanilla & UpLoop & $\Delta$ \\
\midrule
370M & 2.7770 & \textbf{2.7430} & $-0.0340$ \\
760M & 2.5770 & \textbf{2.5381} & $-0.0389$ \\
1.3B & 2.4332 & \textbf{2.3982} & $-0.0349$ \\
\bottomrule
\end{tabular}
\end{small}
\vskip -0.1in
\end{table}

\begin{table}[!t]
\caption{Inference FLOPs per token. UpLoop keeps $L$ independent layers but
executes $L+4$.}
\label{tab:infer}
\vskip 0.1in
\centering
\begin{small}
\begin{tabular}{lccc}
\toprule
Scale & Vanilla & UpLoop & Ratio \\
\midrule
370M & $1.01\times10^9$ & $1.21\times10^9$ & $1.199\times$ \\
760M & $1.92\times10^9$ & $2.32\times10^9$ & $1.210\times$ \\
1.3B & $3.10\times10^9$ & $3.77\times10^9$ & $1.217\times$ \\
\bottomrule
\end{tabular}
\end{small}
\vskip -0.1in
\end{table}

\clearpage
\clearpage
\section{Per-category and Per-task Downstream Results at 370M}
\label{app:main-full}

\begin{table}[t]
\caption{Category means at 370M over eligible tasks only. $\Delta$ is UpLoop $-$ Vanilla. Average equally weights all 13 eligible tasks. Bold marks the better score in each row.}
\label{tab:cat-full}
\vskip 0.1in
\centering
\begin{footnotesize}
\setlength{\tabcolsep}{3pt}
\begin{tabular}{lrrr}
\toprule
Category ($n$ valid) & Vanilla & UpLoop & $\Delta$ \\
\midrule
Closed-book Q\&A (4) & 9.49 & \textbf{9.81} & +0.32 \\
Open-book Q\&A (4) & 18.20 & \textbf{19.51} & +1.32 \\
LAMBADA (1) & 34.02 & \textbf{34.58} & +0.56 \\
HellaSwag (1) & 42.60 & \textbf{44.93} & +2.33 \\
Math word (1) & 1.97 & \textbf{2.52} & +0.55 \\
Reas. primitives (2) & 28.50 & \textbf{36.20} & +7.70 \\
\textbf{Average} (13) & 18.95 & \textbf{20.90} & +1.95 \\
\bottomrule
\end{tabular}
\end{footnotesize}
\vskip -0.1in
\end{table}

\begin{table}[t]
\caption{Per-task downstream results at 370M with equal training tokens and FLOPs, aligned with Table~\ref{tab:main}. $\Delta$ is UpLoop $-$ Vanilla. A dagger marks tasks excluded from all averages; bold marks the better score only for eligible tasks.}
\label{tab:main-full}
\vskip 0.1in
\centering
\begin{footnotesize}
\setlength{\tabcolsep}{4pt}
\begin{tabular}{lrrr}
\toprule
Benchmark & Vanilla & UpLoop & $\Delta$ \\
\midrule
\multicolumn{4}{l}{\textit{Closed-book question answering}} \\
TriviaQA & 13.36 & \textbf{13.72} & +0.36 \\
TyDiQA-NoContext & 8.66 & \textbf{8.81} & +0.15 \\
NaturalQuestions & 5.10 & \textbf{5.53} & +0.43 \\
WebQuestions & 10.85 & \textbf{11.17} & +0.32 \\
\midrule
\multicolumn{4}{l}{\textit{Open-book question answering}} \\
TyDiQA-GoldP & 15.17 & \textbf{16.46} & +1.29 \\
SQuADv2$^{\dagger}$ & 6.35 & 7.01 & +0.66 \\
DROP & 7.09 & \textbf{7.60} & +0.51 \\
QuAC & 23.04 & \textbf{24.88} & +1.84 \\
CoQA & 27.51 & \textbf{29.12} & +1.61 \\
\midrule
\multicolumn{4}{l}{\textit{Language understanding}} \\
LAMBADA & 34.02 & \textbf{34.58} & +0.56 \\
HellaSwag & 42.60 & \textbf{44.93} & +2.33 \\
\midrule
\multicolumn{4}{l}{\textit{Math word problems}} \\
SVAMP$^{\dagger}$ & 1.61 & 2.01 & +0.40 \\
ASDiv & 1.97 & \textbf{2.52} & +0.55 \\
MAWPS AddSub$^{\dagger}$ & 1.39 & 1.39 & +0.00 \\
MAWPS MultiArith$^{\dagger}$ & 2.96 & 4.52 & +1.57 \\
MAWPS SingleOp$^{\dagger}$ & 1.14 & 2.66 & +1.52 \\
MAWPS SingleEq$^{\dagger}$ & 0.83 & 0.83 & +0.00 \\
\midrule
\multicolumn{4}{l}{\textit{Reasoning primitives}} \\
Copying (random words)$^{\dagger}$ & 11.60 & 12.90 & +1.30 \\
Copying (real words)$^{\dagger}$ & 11.30 & 17.00 & +5.70 \\
VarAssign-basic$^{\dagger}$ & 20.30 & 23.90 & +3.60 \\
VarAssign-math & 28.10 & \textbf{34.40} & +6.30 \\
VarAssign-code & 28.90 & \textbf{38.00} & +9.10 \\
\midrule
\textbf{Average} (13 valid of 22) & 18.95 & \textbf{20.90} & +1.95 \\
\bottomrule
\end{tabular}
\end{footnotesize}
\vskip -0.1in
\end{table}

We report category-level and per-task results for the 370M models in Table~\ref{tab:main}, using the same eligible task set as the main text.
Of the 22 tasks, 13 enter Average, yielding scores of 18.95 for Vanilla and 20.90 for UpLoop, an improvement of 1.95 points.
Tables~\ref{tab:cat-full} and~\ref{tab:main-full} report the category means and individual task scores, respectively.
Tasks below the random-answer threshold are reported individually and excluded from category and overall averages.

\section{Downstream Evaluation Details}
\label{app:downstream-eval}

Downstream accuracy in Figures~\ref{fig:scale} and~\ref{fig:tradeoff} is the
unweighted mean over six zero-shot benchmarks: ARC-c, ARC-e, HellaSwag,
PIQA, SIQA and BoolQ. All six models exceed the random-answer threshold on these tasks, determined from 1,000 simulations per task. The same six tasks are used at every scale. All results use final
checkpoints and complete evaluation splits. Values above each pair in
Figure~\ref{fig:scale} report UpLoop minus Vanilla in percentage points.

Held-out NLL in Figure~\ref{fig:tradeoff}(c) is evaluated in FP32 on the same
2,048 held-out sequences using final checkpoints. In
Figure~\ref{fig:tradeoff}(a), open markers indicate the last evaluation before
each stage transition, and dotted segments show the corresponding NLL jumps.
Black dashed segments in Figure~\ref{fig:tradeoff} connect models at the same
scale.

\paragraph{Evaluation implementation.}
Commonsense tasks use lm-evaluation-harness 0.4.9.1.
Checkpoints are converted to the HuggingFace Olmo~2 format, with looping unrolled in its native execution order, and evaluated in FP32.
The conversion reproduces the training-time held-out NLL to within $2\times10^{-4}$.
For the 22-task suite, blank lines are collapsed to single newlines to match the pretraining data format.

\paragraph{Random-answer construction.}
Random-answer thresholds are determined from 1,000 simulations per task.
Answers are sampled uniformly from candidates for multiple-choice tasks, from passage spans for open-book QA, from randomly selected reference answers for closed-book QA and mathematics, and from context words for LAMBADA.
All random answers use the corresponding task scorer.

\paragraph{Main result tables.}
\Cref{tab:commonsense} reports zero-shot accuracy on the seven commonsense
benchmarks used by \citet{he2026onelr}, together with BoolQ.
\Cref{tab:main} reports token F1 for question answering, length-normalized
accuracy for HellaSwag, and accuracy for the remaining tasks, all on a 0--100
scale. Question answering and LAMBADA are zero-shot; mathematics and reasoning
primitives use five examples. VarAssign tasks use depth 0.
At each scale, Average includes all tasks for which both models exceed the
random-answer threshold, based on 1,000 single-answer simulations. A dagger marks pairs excluded from Average;
bold marks the best score within each scale, only for eligible pairs. The valid task counts are
6, 7 and 7 in \cref{tab:commonsense}, and 13, 21 and 22 in
\cref{tab:main}. These averages differ from the fixed six-task average in
Figures~\ref{fig:scale} and~\ref{fig:tradeoff}; averages over different task
sets should not be used to compare scales.

\section{GSM8K Fine-tuning Details}
\label{app:gsm8k-sft}

We fine-tune both 1.3B models on the official 7,473-example GSM8K training
split for three epochs using AdamW, a peak learning rate of $2\times10^{-5}$,
six warmup steps followed by cosine decay, and a global batch size of 128.
Fine-tuning uses a maximum sequence length of 1,024 tokens, with no training examples truncated.
AdamW uses $\beta=(0.9,0.999)$, $\epsilon=10^{-8}$, and zero weight decay.
We supervise only the solution tokens and retain UpLoop's shared-weight
execution structure. Evaluation uses all 1,319 test examples, zero-shot
greedy decoding without a calculator, a 512-token generation limit, and exact
matching of the numerical answer following \texttt{\#\#\#\#}.
The fine-tuning compute is not matched.

Both models exceed the random-answer bound of $1.74\%$, and UpLoop improves
over Vanilla by $3.34$ percentage points. The comparison uses one training
seed (42).

Five-shot evaluation uses five fixed chain-of-thought examples sampled from
the training split with seed 1234, keeping the checkpoints, decoding settings,
and scoring rule unchanged. Both models score lower with five-shot prompting,
while UpLoop retains a $2.65$-point advantage.

\twocolumn[{
\begin{minipage}{\textwidth}
\section{Additional Training-budget Comparisons}
\label{app:training-budget-comparisons}

\captionof{table}{\textbf{Training-budget configurations at 370M.}
Detailed configurations for the completed comparisons in Table~\ref{tab:training-strategies}.
Every row shares \textsc{Vanilla}'s width, tokenizer, ordered data stream, global batch, peak learning rate, and seed (42); cosine decay spans each run's own token budget.
Loop with dense initialization first trains Dense-16 for 20,919 updates, followed by 26,290 updates with looping; the start and duration of the looping stage match UpLoop.
$N$ counts non-embedding parameters, and $D_0$ and $C_0$ are the dense baseline's token and analytical training-FLOP budgets.
LIDAS extends the data stream to match $C_0$ within $0.00066\%$ through whole-update rounding.}
\label{tab:baselines}
\vskip 0.1in
\centering
\begin{small}
\setlength{\tabcolsep}{5pt}
\begin{tabular}{llcrccc}
\toprule
Pays with & Model & Layers (unique/exec.) & $N$ & $D/D_0$ & $C/C_0$ & NLL $\downarrow$ \\
\midrule
--- & \textsc{Vanilla}, dense & 16 / 16 & 371M & 1.000 & 1.000 & 2.7770 \\
\midrule
\multirow{3}{*}{training FLOPs}
 & Loop from scratch, $\mathrm{Loop}(6\text{-}4{\times}2\text{-}6)$ & 16 / 20 & 371M & 1.000 & 1.199 & 2.7587 \\
 & Dense-20 (Iso-Tokens)                                             & 20 / 20 & 438M & 1.000 & 1.199 & 2.7390 \\
 & Loop with dense initialization                                  & 16 / 20 & 371M & 1.000 & 1.111 & 2.7587 \\
\midrule
tokens
 & Dense-20 (Iso-FLOPs)                                              & 20 / 20 & 438M & 0.834 & 1.000 & 2.7608 \\
\midrule
extra tokens
 & LIDAS (Iso-FLOPs)~\citep{kapl2025depth}                           & 16 / 16 & 371M & 1.204 & 1.000 & 2.7514 \\
\midrule
early depth
 & \textbf{UpLoop (ours)}: grow, then loop the block                 & 16 / 20 & 371M & \textbf{1.000} & \textbf{1.000} & \textbf{2.7430} \\
\bottomrule
\end{tabular}
\end{small}
\vskip -0.1in
\end{minipage}
\vspace{0.2in}
}]

\raggedbottom
\section{Additional Mechanistic Analysis}
\label{app:additional-mechanism}
We further analyze the final pretrained models through cross-call output substitution, attention-component interventions, and task-specific information reading. The first two analyses compare the 370M UpLoop, Loop with dense initialization, and Loop from scratch models, which share the same final looped architecture. The contextual retrieval analysis uses 760M and 1.3B UpLoop and Vanilla models. Experimental details are provided in Appendix~\ref{app:analysis-details}.

\subsection{Cross-call Output Compatibility}
\label{app:cross-call-compatibility}
Shared parameters process different hidden states across calls, motivating us to examine whether updates from the first call remain usable in the second. In each intervention, we replace the attention or MLP update at one second-call core layer with the update from its first call, match the native second-call norm tokenwise, and recompute the remaining network. Table~\ref{tab:analysis-cross-call} reports the mean $\Delta\mathrm{NLL}$ across separate substitutions at the four core positions.

UpLoop incurs substitution costs of 0.00919 for attention and 0.01929 for MLP, both lower than those of the two looping baselines. These results indicate greater functional compatibility between sublayer updates produced across calls in UpLoop.

\begin{table}[!ht]
\caption{Mean $\Delta\mathrm{NLL}$ under cross-call update substitution. Second-call updates are replaced with first-call updates from the same shared layer after tokenwise matching to native second-call norms. Costs are averaged over four separate single-position substitutions within each window, then over 512 windows. Each window contains 512 tokens, with the final 128 scored.}
\label{tab:analysis-cross-call}
\centering\small
\begin{tabular}{p{.51\columnwidth}rr}
\toprule
Model & Attention & MLP \\
\midrule
UpLoop & \textbf{0.00919} & \textbf{0.01929} \\
Loop with dense initialization & 0.01427 & 0.03426 \\
Loop from scratch & 0.02840 & 0.05341 \\
\bottomrule
\end{tabular}
\end{table}

\subsection{Compatibility and Difference Propagation along the Value Path}
\label{app:value-path}
\subsubsection{Attention Weights and Values}
\begin{figure}[!ht]
\centering
\includegraphics[width=\columnwidth]{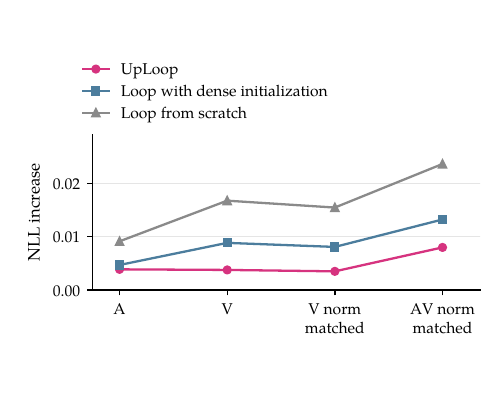}
\caption{Cross-call substitution of attention weights and values. A and V reuse first-call attention weights and values, respectively. V norm matched rescales reused values to current norms per token and head; AV norm matched combines reused attention weights with norm-matched values. Each condition averages separate interventions at the four core positions over 64 windows.}
\label{fig:analysis-av-reuse}
\end{figure}

To localize the components underlying attention-output compatibility, we separately reuse first-call attention weights and value vectors, and additionally norm-match the reused values. Figure~\ref{fig:analysis-av-reuse} shows that norm-matched value reuse incurs a cost of 0.00352 in UpLoop, compared with 0.00809 for Loop with dense initialization and 0.01546 for Loop from scratch. Reusing attention weights alone yields a less distinct difference between UpLoop and the former baseline. Values thus provide a clearer component-level instance of cross-call compatibility, with their lower substitution cost persisting after amplitude control.

\subsubsection{Propagation of Substitution-induced Differences}
\begin{figure}[H]
\centering
\includegraphics[width=\columnwidth]{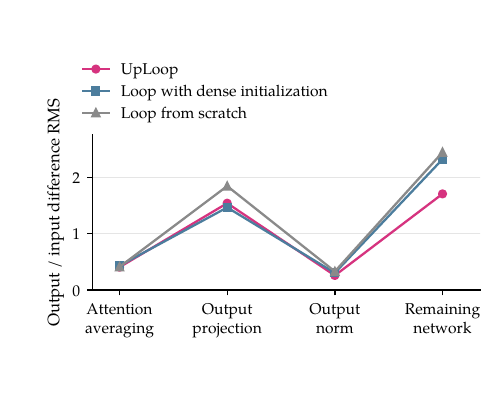}
\caption{Stagewise transmission of differences induced by old-value substitution. Each point is the ratio of output-difference RMS to input-difference RMS for the corresponding stage. The remaining-network term measures transmission from changes in normalized attention updates to changes in final hidden states. Ratios are averaged across the four core positions within each window, then across 64 windows.}
\label{fig:analysis-value-transport}
\end{figure}

We next trace how differences introduced by old-value substitution propagate through the forward path. Keeping native Q/K at the intervention site, we measure successive RMS ratios through attention averaging, output projection, output normalization, and the remaining network. Figure~\ref{fig:analysis-value-transport} shows an output-normalization gain of 0.261 for UpLoop, compared with 0.306 and 0.332 for the two baselines. From the normalized attention update to the final state, the respective gains are 1.708, 2.322, and 2.445.

These measurements locate lower difference transmission at output normalization and subsequent propagation: along the changes induced by actual old-value substitution, UpLoop attenuates more at normalization and amplifies less through the remaining network. Output projection exhibits a different ordering across models, revealing distinct responses to substitution-induced differences across stages.

\subsection{Additional Evidence for Contextual Retrieval}
\label{app:contextual-retrieval}
\subsubsection{Answers under Changed Variable Bindings}
\label{app:binding-swaps}
Accurate contextual retrieval also requires answers to follow changes in variable bindings. We swap the values assigned to the target variable and another variable while keeping the query, variable set, and value set fixed, and measure the fraction of pairs answered correctly both before and after the swap. Table~\ref{tab:analysis-binding} shows that UpLoop achieves average paired accuracies of 77.7\% and 83.2\% at 760M and 1.3B, compared with 42.6\% and 45.7\% for Vanilla, indicating that its answers more reliably follow the current variable--value bindings.

\begin{table}[ht]
\caption{Paired accuracy (\%): the fraction of pairs answered correctly both before and after the swap. Each task has 128 pairs. Values are exchanged between the queried variable and the next assignment, wrapping from last to first, with the query, other assignments, demonstrations, and candidate set fixed. Average equally weights all eligible tasks, Math and Code.}
\label{tab:analysis-binding}
\vskip 0.1in
\centering
\begin{small}
\begin{tabular}{llrr}
\toprule
Scale & Task & Vanilla & UpLoop \\
\midrule
760M & Math & 39.8 & \textbf{74.2} \\
     & Code & 45.3 & \textbf{81.2} \\
     & Average & 42.6 & \textbf{77.7} \\
\midrule
1.3B & Math & 34.4 & \textbf{81.2} \\
     & Code & 57.0 & \textbf{85.2} \\
     & Average & 45.7 & \textbf{83.2} \\
\bottomrule
\end{tabular}
\end{small}
\vskip -0.1in
\end{table}
\subsubsection{Target Reading across Core Calls}
\label{app:reading-stages}
Table~\ref{tab:analysis-core-reading} compares target-value reading between the two calls to the shared core in 1.3B UpLoop, blocking access in all four core layers within each call. The analysis uses the same 128 original examples per task as the swap experiment. Relative to blocking an unrelated value, target blocking yields an average accuracy loss of 0.0 percentage points in the first call and 14.5 points in the second, with second-call losses of 14.1 and 14.8 points on Math and Code, respectively. These results show that the second call to the shared core supports answer formation through target-information reading.

\begin{table}[!ht]
\caption{Target reading across the two core calls in 1.3B UpLoop. Each condition blocks and renormalizes attention from the query suffix to the target-value token or an unrelated value token in all four core layers of the corresponding call. Values report control-block minus target-block accuracy in percentage points. Each task contains 128 examples. Average equally weights all eligible tasks, Math and Code.}
\label{tab:analysis-core-reading}
\centering\small
\begin{tabular}{lrr}
\toprule
Task & Core 1 (pp) & Core 2 (pp) \\
\midrule
Math & 0.0 & 14.1 \\
Code & 0.0 & 14.8 \\
Average & 0.0 & 14.5 \\
\bottomrule
\end{tabular}
\end{table}

\clearpage
\section{Further Analysis: Experimental Details}
\label{app:analysis-details}

\paragraph{Gradient statistics.}
For Figure~\ref{fig:analysis-gradient}, we unroll shared calls into independent copies with identical weights at the final checkpoint and backpropagate the prediction loss to obtain call-specific gradient contributions.
Cosine similarity uses gradients over all parameters within each layer, averaged first over the four shared layers and then over 12 validation windows of 256 tokens each.

\paragraph{State perturbations.}
For Table~\ref{tab:analysis-perturbation}, we inject a single random-direction perturbation per token after the first core pass, with magnitude equal to 1\% of its pre-core state norm. Trajectory distance is the mean tokenwise L2 distance between the perturbed and unperturbed states, normalized by the distance immediately after perturbation. Results average 512 windows.

\paragraph{Layer substitution.}
For Figure~\ref{fig:analysis-lineage}, targets are the four first-pass core layers.
Replacement layers are grouped by initial ancestry traced through the full growth history, excluding the target itself.
Each replacement layer processes the target layer's input, and its attention or MLP update replaces the original after tokenwise norm matching.
Each of the 64 windows contains 256 tokens, with the final 64 scored.
Substitution costs are averaged within each ancestry group per window, then across windows.

\paragraph{VarAssign position groups.}
Position analysis in Figure~\ref{fig:analysis-position} uses all 1,000 examples per task from the main evaluation, grouping targets in the first four assignments and the final assignment.
Each group uses the same random-answer criterion as the main text, with eligible tasks weighted equally.
Math and Code are eligible in every group and both enter the average.
Reading blocks and binding swaps use 128 examples per task.

\paragraph{Reading blocks.}
For Table~\ref{tab:analysis-reading}, attention from all tokens at and after the query start to the target-value token is blocked in all four core layers during the second call and renormalized over the remaining positions. The control blocks an unrelated value token in the same example. Each task uses the same 128 original examples as the variable-binding swap experiment in Appendix~\ref{app:binding-swaps}.

\end{document}